# A novel method for predicting and mapping the presence of sun glare using Google Street View

Xiaojiang Li [a], Bill Yang Cai [a], Waishan Qiu [a], Jinhua Zhao [b], Carlo Ratti [a]

[a] *MIT Senseable City lab, Department of Urban Studies and Planning, Massachusetts Institute of Technology, Room 9-250, 77 Massachusetts Avenue, Cambridge, MA, 02139, United States, USA*

[b] *Department of Urban Studies and Planning, Massachusetts Institute of Technology, Cambridge, MA 02139, United States*

Abstract: The sun glare is one of the major environmental hazards that cause traffic accidents. Every year, many people died and injured in traffic accidents related to sun glare. Providing accurate information about when and where sun glare happens would be helpful to prevent sun glare caused traffic accidents and save lives. In this study, we proposed to use publicly accessible Google Street View (GSV) panorama images to estimate and predict the occurrence of sun glare. GSV images have view sight similar to drivers, which would make GSV images suitable for estimating the visibility of sun glare to drivers. A recently developed convolutional neural network algorithm was used to segment GSV images and predict obstructions on sun glare. Based on the predicted obstructions for given locations, we further estimated the time windows of sun glare by estimating the sun positions and the relative angles between drivers and the sun for those locations. We conducted a case study in Cambridge, Massachusetts, USA. Results show that the method can predict the presence of sun glare precisely. The proposed method would provide an important tool for drivers and traffic planners to mitigate the sun glare and decrease the potential traffic accidents caused by the sun glare.

## 1. Introduction

The sun glare is one of the major safety hazards that cause traffic accident for drivers (Hagita and Mori, 2014; Jurado-Piña, 2010). When the sun is low during rising and setting and the sunlight may enter the field of view of drivers at some time with the right azimuth angles. The glare caused by sun would create a risk to drivers. The sun glare would impair driver's vision and cause temporarily blindness to drivers, both of which are annoying and dangerous for drivers (Aune, 2017). Hagita and Mori (2014) investigated the effects of sun glare on traffic accidents. Results show that sun glare is connected with traffic accidents of cars and cyclists. Hagita and Mori (2013) examined the association between sun glare and bicycle and pedestrian accidents in Chiba prefecture, Japan. Results show that sun glare increases traffic accidents in cases of oncoming bicycles and pedestrians encountering turning vehicles significantly. Sun glare would also slow down traffic flows (Andrew et al., 2012; Auffray et al., 2008). Auffray et al (2008) found that the presence of sun glare would affect the traffic flow during both congested and uncongested periods. Andrew et al (2012) studied the connection between the sun glare and the freeway congestion, and result shows that the presence of sun glare would slow traffics and cause traffic congestions.

Visor and tinting are very commonly used method to avoid sun glare. However, the visor would obstruct driver's overall vision, while driving without obstruction is one of the basic requirements for safe driving. In addition, the visor and the tinting can only cover small portion of the front windshield and cannot screen out the sun (Hagita and Mori, 2013). The visor would also block the driver's view sight to see the overhead traffic light. For drivers, knowing when and where the sun glare occurs would help to take some measures to avoid the potential sun glare caused traffic accidents and congestions. For road builders, knowing the spatio-temporal distributions of sun glare would help to design safer road networks and provide better driving experiences to drivers with minimum sun glare. Accurate prediction of the presence of the sun glare is thus needed.

The occurrence of sun glares is determined by sun's angle in relation to cars, which is further influenced by the sun elevation, azimuth, driving direction, and slope. The terrains changes and obstructions along roads could block the sunlight from shining to driver's eyes. Currently, using the digital terrain model and the road network map is the most widely used method to predict the presence of the sun glare (Hagita and Mori, 2014; Kjetil et al., 2017). The sun position at a specific time of the vehicle's relative location can be precisely estimated. Based on the geometrical model between the driving car and the sun position, the occurrence of the sun glare can be predicted (Jurado-Piña and Pardillo-Mayora, 2010). While the digital terrain model considers of the obstruction effect of sun glare caused by the terrain, however, the obstruction of the building blocks and trees on both sides of roads are not considered in the digital terrain based method. In addition, the high spatial resolution digital terrain model is usually required for accurate sun glare prediction. However, high spatial resolution digital terrain models usually are not available.

In this study, we proposed to use the Google Street View (GSV) panorama images to predict the occurrence of sun glare. Since GSV images were taken by moving vehicles on the road (Li et al., 2015), those street-level images would perfectly represent driver's visual experience in driving. In addition, GSV images capture the obstructions along roads at a fine level, which would further provide accurate information of obstructions along roads. We used the state-of-the-art deep convolutional neural network algorithm to segment GSV images and classify those obstruction pixels from GSV images. Based on the image segmentation results and the geometrical model of the sun and drivers, we predicted the presence of sun glare with consideration of all obstructions such as, terrain changes, trees, buildings, etc. This study would provide a very cheap and efficient way to predict the spatio-temporal distribution of the sun glare at a large scale precisely.

## 2. Methodology

2.1 Study area and dataset preparation

City of Cambridge, Massachusetts was chosen as the case study area. In order get fine level sun glare maps, we first created sample sites along streets every 40 meters in the study area. Those sample sites were used to collect Google Street View (GSV) panorama images and visualize the sun glare presence. **Fig. 1** (a) shows the generated sample sites along streets in Cambridge, Massachusetts. Based on the coordinates of these created sample sites, we further collected GSV panoramas from Google Server using Google Maps Application Programming Interfaces (APIs). In this study, we developed a Python script (**Appendix A**) to download GSV panoramas for all sites using the coordinates of those sample sites as the input.

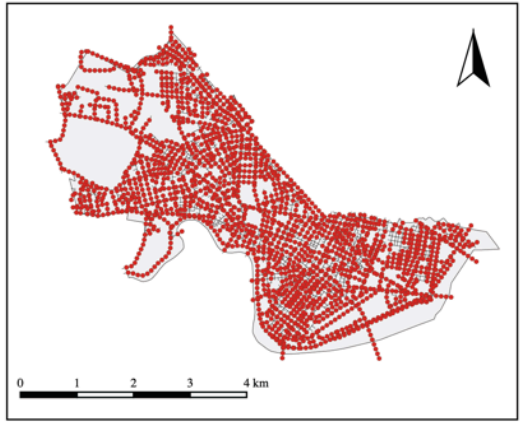

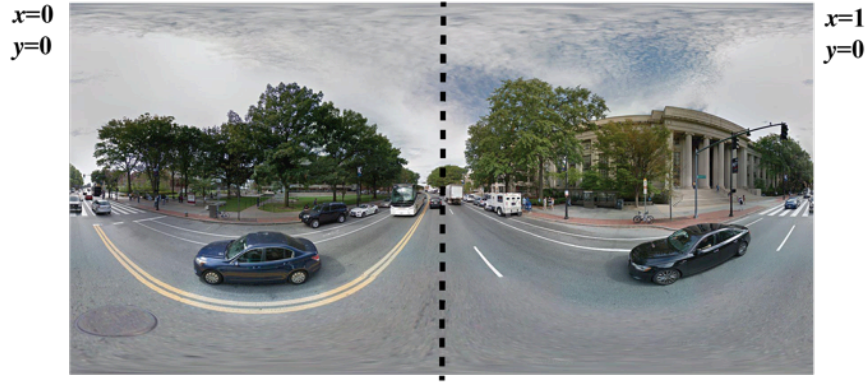

**Fig. 1.** The workflow for GSV panorama collection in Cambridge, (a) the created sample sites (b) the metadata of GSV panoramas, (c) a GSV panorama of one sample site.

2.2 Google Street View (GSV) panoramas segmentation

Recent progresses in artificial intelligence make it possible to recognize obstructions along streets from GSV panoramas accurately. In this study, we applied the Pyramid Scene Parsing Network (PSPNet), a deep convolutional neural network trained on the ADE20K dataset (Zhao et al., 2017; Zhou et al., 2017) to segment GSV panoramas. The PSPNet achieved the state-of-the-art performance in pixel-level segmentation of complex visual scenes in the ADE20K dataset.

Hence, the validated performance of the PSPNet trained on the ADE20K dataset gives us confidence in our use of the PSPNet to accurately classify GSV panoramas into sky, buildings, trees, ground, and other environmental factors. **Fig. 2** shows the image segmentation results on sample GSV images and panoramas using the PSPNet trained on ADE20K dataset. Those non-sky pixels were considered as obstructions, which would block the glaring sun.

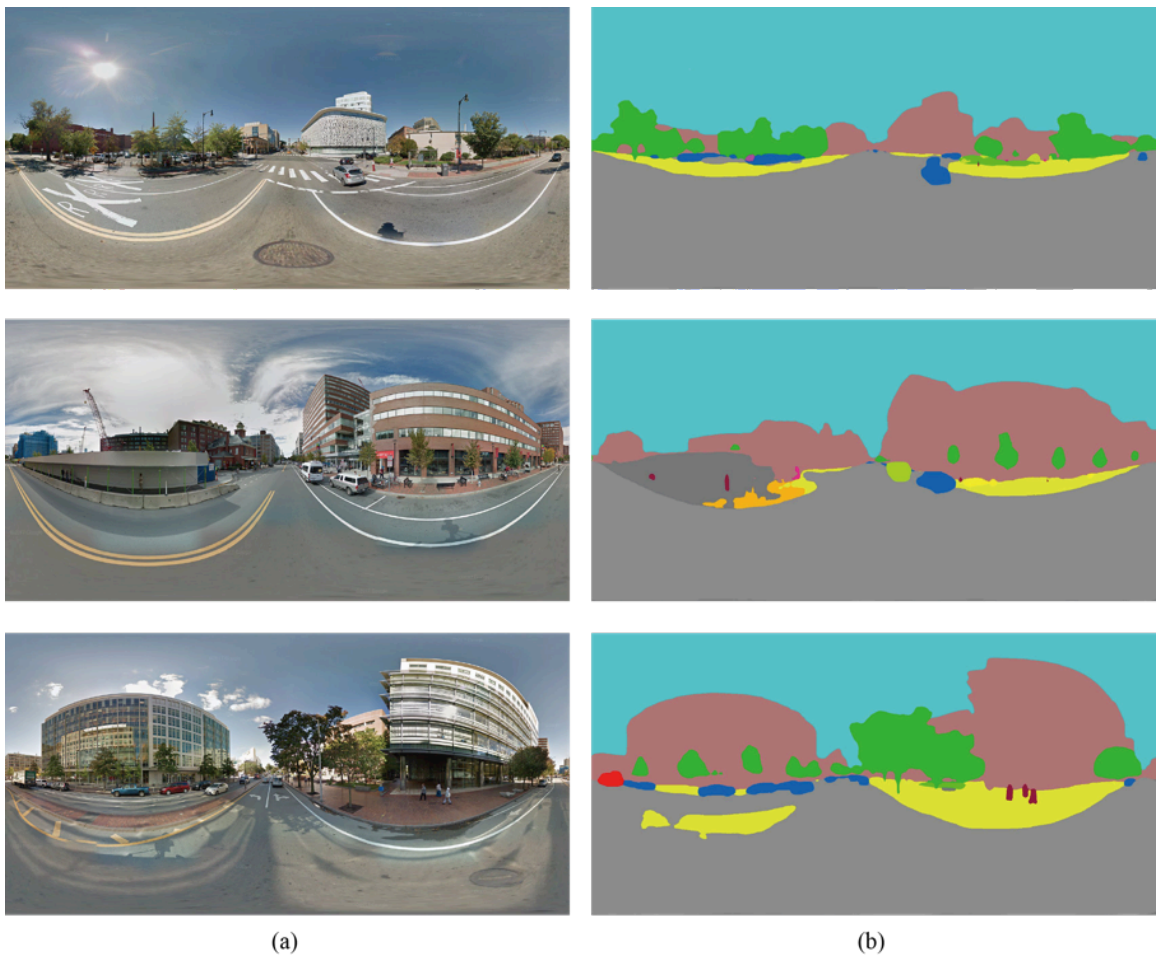

**Fig. 2.** Image segmentation results on GSV panoramas using pyramid scene parsing convolutional neural network (PSPnet).

2.3 Sun glare prediction

The position of sun (sun elevation $\theta$ and azimuth angle $\phi$) can be estimated accurately for any location (*lon*, *lat*) at one specific time (*t*).

$$\begin{aligned} \theta &= f(lon, lat, t) \\ \phi &= g(lon, lat, t) \end{aligned} \qquad (1)$$

In this study, we used an open source python module "*pysolar*" to estimate the sun position for any location at a specific time.

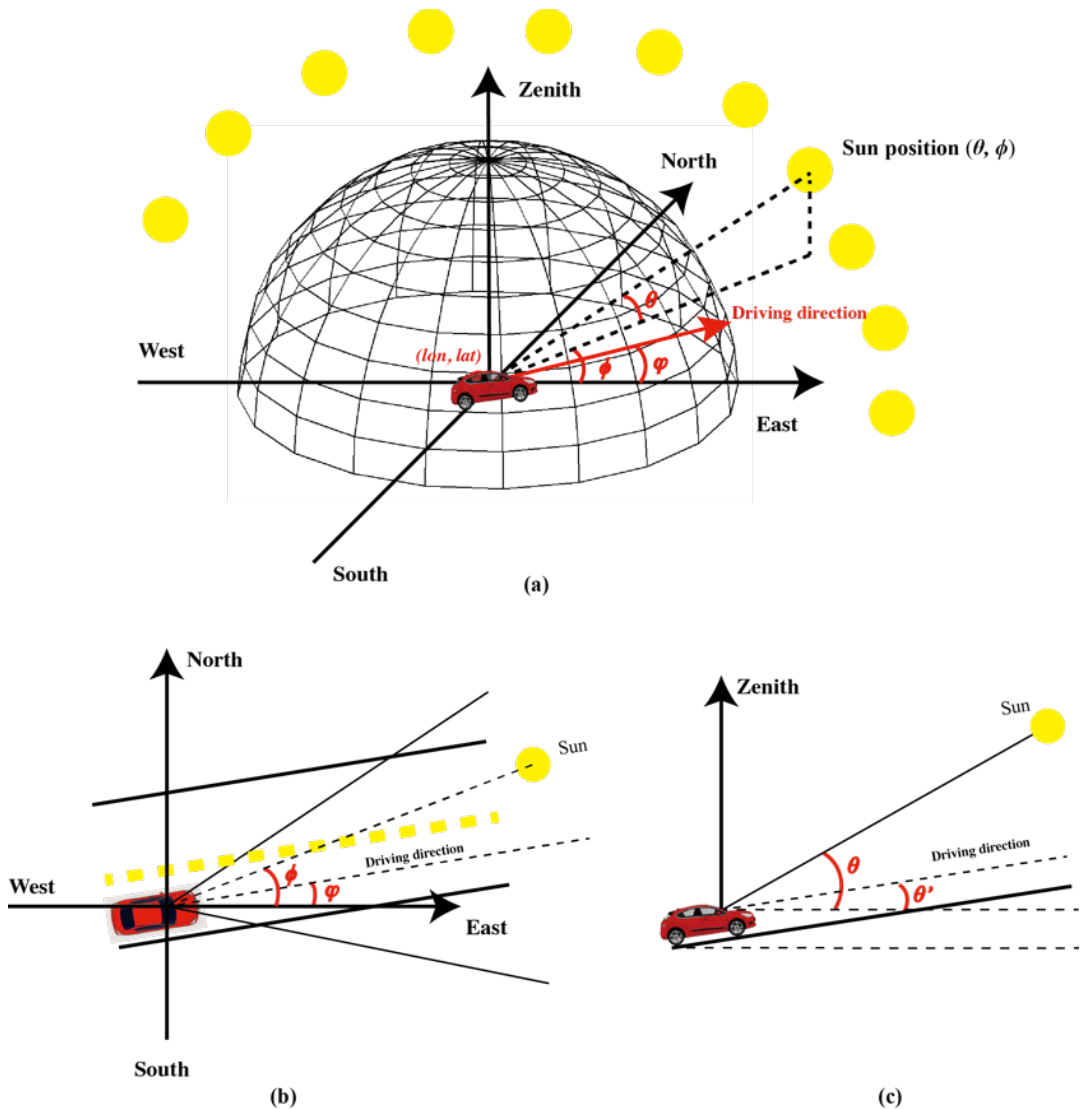

**Fig. 3.** The geometrical model of the sun position and the location of vehicle on driveway.

Fig. 3 (a) shows the geometrical model of the sun position and the location of a car driving through point (*lon*, *lat*), therefore, the relative horizontal direction between sun and the car is (**Fig. 3**(b)),

$$h\_glare = |\phi - \varphi| \quad (2)$$

where the $\phi$ is the azimuth angle of the sun. The relative vertical angle of the car and sun is (**Fig. 3** (c)),

$$v\_glare = |\theta - \theta'| \quad (3)$$

where $\theta'$ is the slope angle of the driveway. The driving direction and the slope angle of the driveway can be accessed from the GSV metadata. Based on previous studies (Jurado-Piña and Mayora, 2009b; Aune, 2017), if the *h_glare* and *v_glare* both are lower than 25 degree and there is no obstruction of the sun light, sun would cause glare to drivers.

The existence of the obstructions on both sides of the driveways could block the sun glare. In this study, GSV panoramas were used to judge if the sun glare is obstructed or not based on the segmented GSV panoramas. **Fig. 4** shows the geometrical model of projecting sun positions on GSV panoramas. The incidence point located on the cylindrical GSV panorama should be ($x_c$, $y_c$),

$$x_c = \frac{\varphi - \phi}{2\pi} W_c + C_x$$
$$y_c = C_y - \frac{\theta - \theta'}{r_0} H_c \quad (4)$$

where $\phi$ is the azimuth angle of the sun, $\varphi$ is the driving direction, $\theta$ is the sun elevation, and $\theta'$ is the tilt angle of GSV panorama, and the ($C_x$, $C_y$) *is* the central pixel of the GSV panorama,

$$C_x = \frac{W_c}{2}$$
$$C_y = \frac{H_c}{2}$$
(5)

Whether the sun is obstructed or not can be estimated by judging whether the incidence point on the GSV panorama is open sky pixel or non-sky obstruction pixel. If the incidence point on the GSV panorama is non-sky pixels, therefore, the sun is obstructed, and the driver at this location of this time has no sun glare problem.

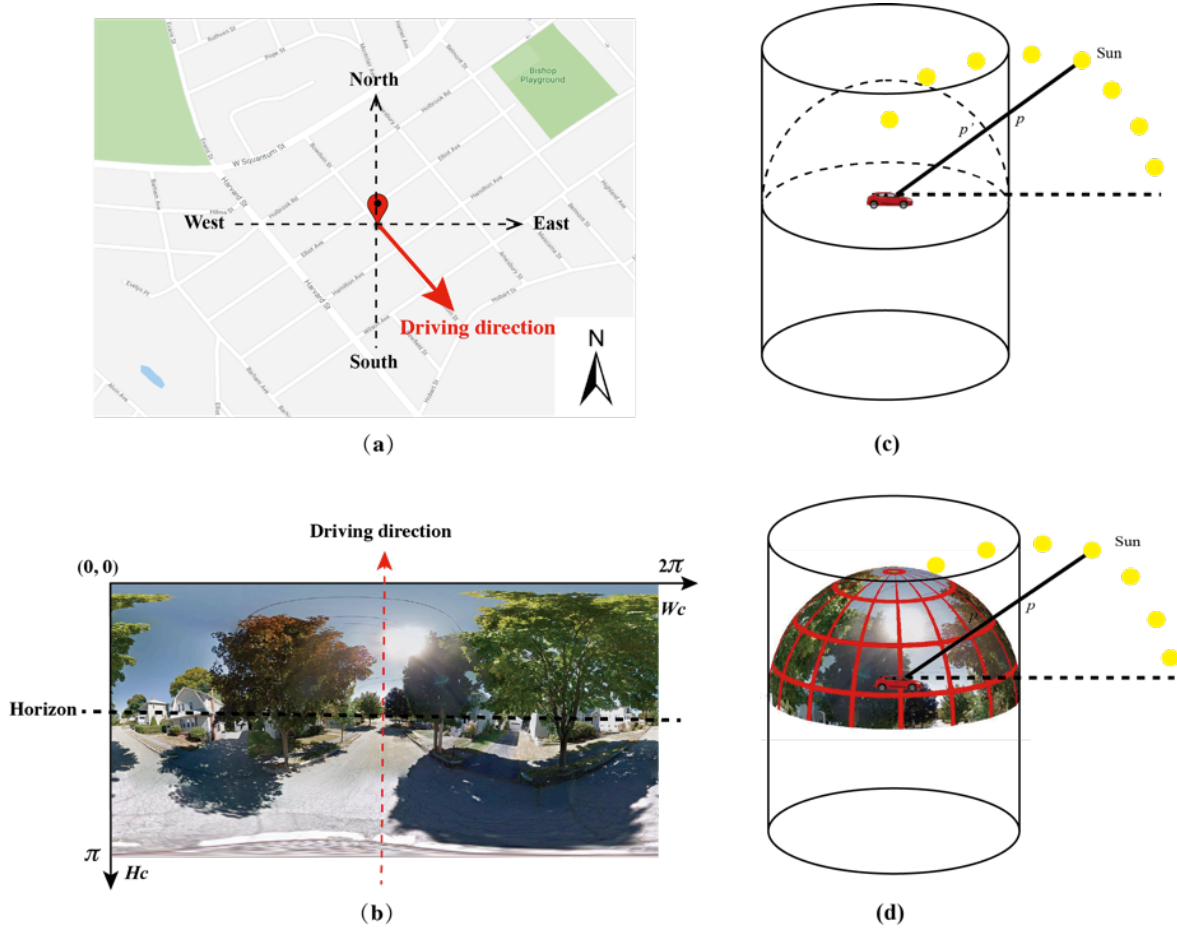

**Fig. 4.** The geometrical model of projecting sun position at a certain time on a cylindrical GSV panorama.

## 3. Results

The sun glare usually happens during sun rising and sun setting when the sun is very low in the sky. **Fig. 5** shows the profiles of sun elevation and azimuth angles from 5 am to 8 pm on 20[th] days of different months in one year at Cambridge, Massachusetts. In Cambridge, June has the highest sun in one year, and December has the lowest sun in one year (**Fig. 5**(a-b)).

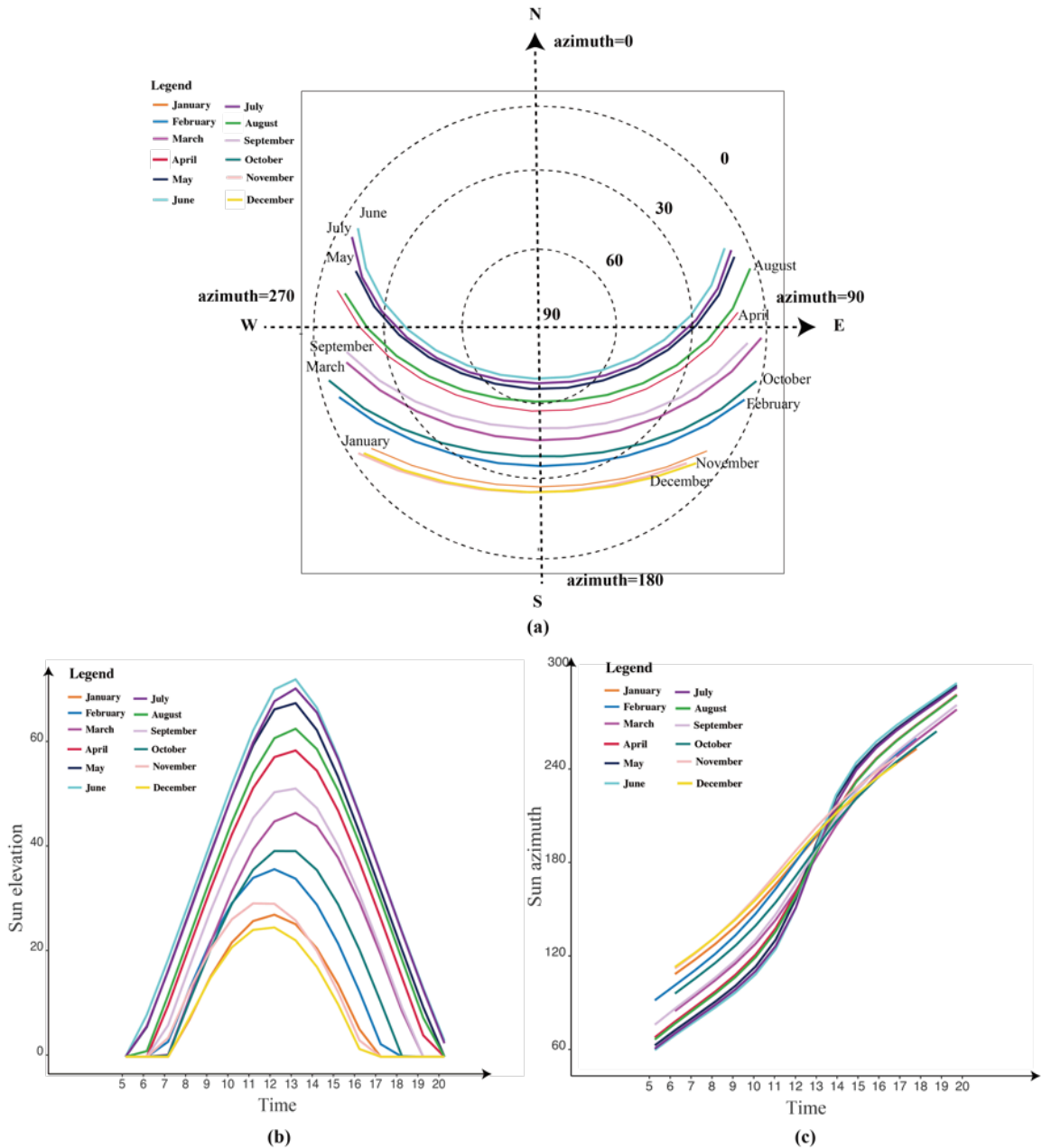

**Fig. 5.** The sun elevation and sun azimuth angles from 5am to 8pm on 20[th] day of different months in Cambridge, MA, (a) the sun diagram of Cambridge in one year, (b) the profiles of the sun elevation angles from 5 am to 8pm on 20$^{th}$ day of different months, (c) the profiles of sun azimuth angles in one day from 5 am to 8pm on 20$^{th}$ day of different months.

If obstructions are not considered, sun glare happens when the relative horizontal direction and the relative vertical angle between sun and the car both are less than 25 degree (**Fig. 3**(b), (c)). The presence of sun glare is influenced by the locations and the orientations of streets. Those streets on the direction of the southeast and southwest have larger possibility to have sun glare in one day. In summer, those streets toward the northeast and northwest will also have the possibility of having sun glare at morning and afternoon, respectively. In winter, the sun position is more southward, and the streets on the direction of the northeast and northwest have no sunrise and sunset sun glare. **Table 1** presents the time at hour level and ranges of the street orientations for sun glare presence for 20$^{th}$ day in different months of 2018 is the obstruction along streets is not considered. Under no obstruction situation, the street orientation ranges for sun glare presence shift in different months of one year. Generally, in winter (e.g. December 20$^{th}$, January 20$^{th}$), driving toward northeast and northwest will have no risk of meeting sun glare in one day. The sunrise sun glare and sunset sun glare happen while driving toward southeast and southwest, respectively. In summer (e.g. June 20$^{th}$, July 20$^{th}$), because sun is highest in one year, driving toward northeast and northwest will have the sunrise and sunset sun glare, respectively.

**Table 1**. The ranges of street orientations with sunrise and sunset sun glare at Cambridge, MA (42.376, -71.117) based on the 25-degree criteria.

| Date, 2018 | Sunrise sun glare (time, ranges) | Sunset sun glare (time, ranges) |
|---|---|---|
| **January 20$^{th}$** | 8 am, [100.87, 150.87] | 2 pm, [186.47, 236.47] |
| | 9 am, [112.51, 162.51] | 3 pm, [199.44, 249.44] |
| | 10 am, [125.72, 175.72] | 4 pm, [210.87, 260.87] |
| **February 20th** | 7 am, [83.46, 133.46] | 3 pm, [204.39, 254.39] |
| | 8 am, [94.29, 144.29] | 4 pm, [216.58, 266.58] |
| | 9 am, [106.58, 156.58] | 5 pm, [227.36, 277.36] |
| **March 20th** | 8 am, [76.74, 126.74] | 5 pm, [225.43, 275.43] |
| | 9 am, [87.93, 137.93] | 6 pm, [236.31, 286.31] |
| **April 20th** | 7 am, [59.16, 109.16] | 6 pm, [246.62, 296.62] |
| | 8 am, [69.24, 119.24] | 7 pm, [256.53, 306.53] |
| **May 20th** | 6 am, [43.47, 93.47] | 6 pm, [253.74, 303.74] |
| | 7 am, [52.95, 102.95] | 7 pm, [263.17, 313.17] |
| **June 20th** | 6 am, [40.15, 90.15] | 6 pm, [255.77, 305.77] |
| | 7 am, [49.40, 99.40] | 7 pm, [264.92, 314.92] |
| | | 8 pm, [274.40, 324.40] |
| **July 20th** | 6 am, [41.43, 91.43] | 6 pm, [252.54, 302.54] |
| | 7 am, [50.93, 100.93] | 7 pm, [261.93, 311.93] |
| **August 20th** | 7 am, [57.82, 107.82] | 6 pm, [246.25, 296.25] |
| | 8 am, [67.84, 117.84] | 7 pm, [256.14, 306.14] |
| **September 20th** | 7 am, [68.01, 118.01] | 5 pm, [228.65, 278.65] |
| | 8 am, [78.45, 128.45] | 6 pm, [239.27, 289.27] |
| **October, 20th** | 8 am, [88.26, 138.26] | 4 pm, [210.60, 260.60] |
| | 9 am, [99.75 149.75] | 5 pm, [222.02, 272.02] |
| **November 20th** | 7 am, [94.54, 144.54] | 1 pm, [178.34, 228.34] |
| | 8 am, [105.44, 155.44] | 2 pm, [192.28, 242.28] |
| | 9 am, [117.76, 167.76] | 3 pm, [204.57, 254.57] |
| | 10 am, [131.72, 181.32] | 4 pm, [215.45, 265.45] |
| **December 20th** | 8 am, [105.45, 155.45] | 12 pm, [159.55, 209.55] |
| | 9 am, [117.08, 167.08] | 1 pm, [174.28, 224.28] |
| | 10 am, [130.15, 180.15] | 2 pm, [187.92, 237.92] |
| | 11 am, [144.52, 194.52] | 3 pm, [200.11, 250.11] |

Based on the ranges of street orientation (**Table 1**), those streets have sunrise glare and sunset glare on Dec 20$^{th}$ have orientations in range of 105 to 195 and 159 to 251, respectively. This can be further confirmed by the spatial patterns of the sunrise sun glare (**Fig. 6**(a)) and sunset sun glare (**Fig. 6**(b) on Dec 20$^{th}$. Those streets orient to southeast have sunrise sun glare, those streets orient southwest have sunset sun glare. On June 20$^{th}$, streets with orientation in ranges of 40 to 60 and 255 to 324 have sunrise and sunset sun glares, respectively. **Fig. 6**(c, d) shows the spatial distributions of the sunrise and sunset glares on June 20$^{th}$, respectively. Those streets orient to southeast have sunrise sun glare, those streets orient southwest have sunset sun glare.

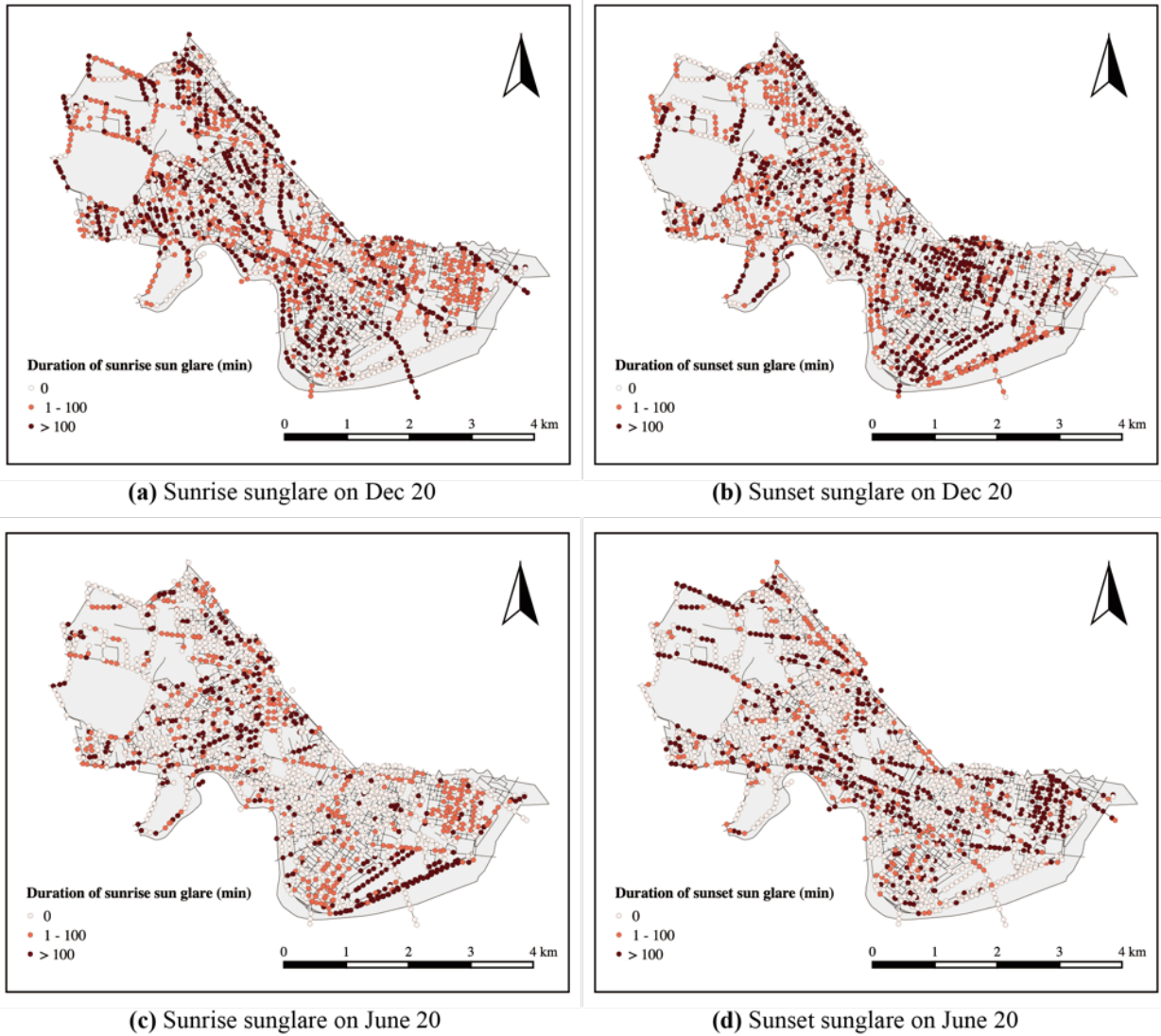

**Fig. 6.** The spatial distributions of the sunrise and sunset sun glares without considering the obstruction on sides of roads, (a) the spatial distribution of sunrise sun glare on December 20$^{th}$, (b) the spatial distribution of the sunset sun glare on December 20$^{th}$, (c) the spatial distribution of sunrise sun glare on June 20$^{th}$, (d) the spatial distribution of sunset sun glare on June 20$^{th}$.

The actual presence of the sun glare map is not only influenced by the orientation of the streets, but also the existence of obstructions between the driving car and the solar position. In order to map the actual sun glare map, we collected Google Street View (GSV) metadata that records the time stamps of all historical GSV panoramas. We categorized all GSV panoramas by captured date into leaf-on season (May, June, July, August, September, and October) and leaf-off season (November, December, January, February, March, and April). **Fig. 7** (a), (b) show the spatial distributions of available GSV panoramas that were captured in leaf-on season and leaf-off season, respectively. Based on these spatial distributions, we found that most the sample sites have GSV panoramas captured in summer available (**Fig. 7**(a)). However, there are only 400 sites having GSV panoramas captured in winter (**Fig. 7**(b)). These four hundred sites would not be suitable to map the whole spatial distribution of the sun glare in winter in the study area. Therefore, it is not suitable to generate sun glare maps for winter based on GSV images currently in the study area. With more street-level images available in winter in future, it would be possible to map the spatio-temporal distributions of the sun glare map. In this study, we only present the sun glare map in leaf-on season.

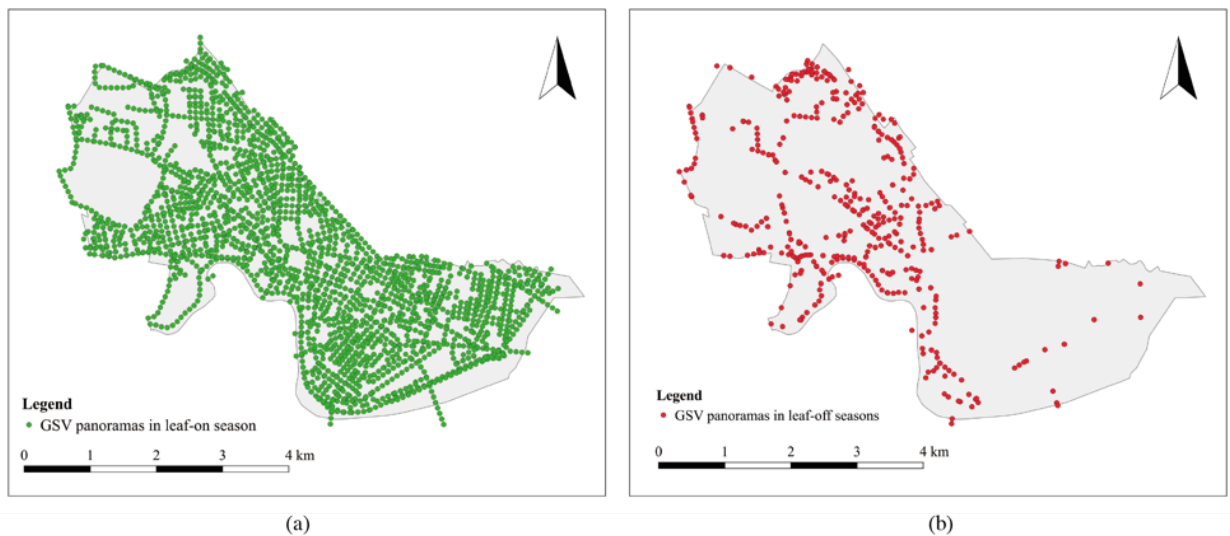

**Fig. 7.** The spatial distributions of available GSV panoramas in leaf-on season (a) and leaf-off season (b).

Based on the sun elevation and the azimuth, the sun path can be projected on the cylindrical GSV panorama. **Fig. 8** shows the sun paths at different hours in days of July 15$^{th}$, August 15$^{th}$, September 15$^{th}$, and October 15$^{th}$ in 2018. We assumed the existence of the buildings and trees would block the sun glare. Therefore, sun glare would be obstructed if sun position were not on sky pixels.

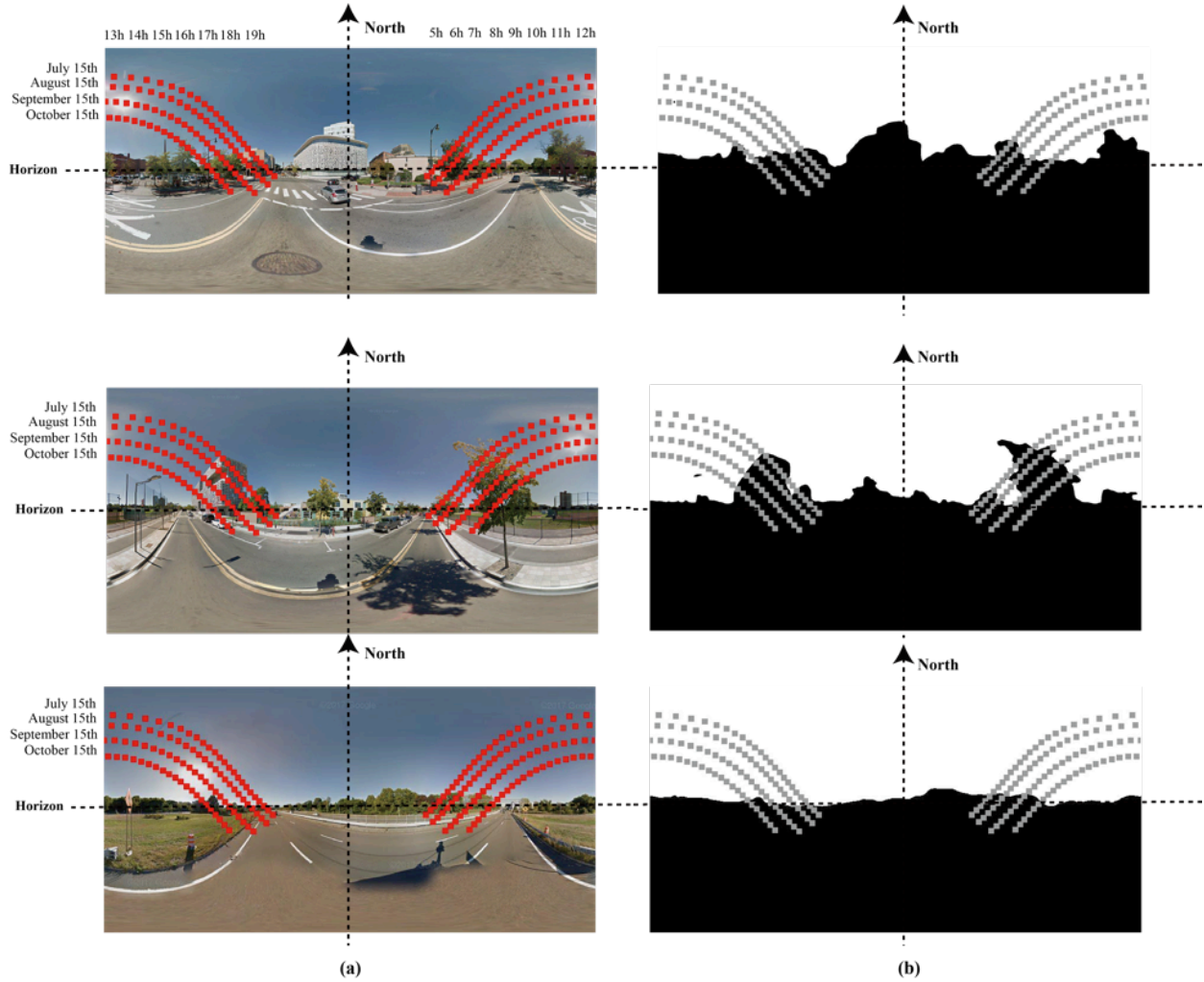

**Fig. 8.** The sun paths in July 15$^{th}$, August 15$^{th}$, September 15$^{th}$, and October 15$^{th}$ on cylindrical Google Street View panoramas (a) and the segmented cylindrical panoramas (b) at three sites of the study area.

In order to validate the proposed method, we further took *in situ* measurement in the study area. **Fig. 9** shows the comparison of the predicted sun path of July 5$^{th}$ and actual sun positions on photos taken at the same location. The predicted result shows that the sun is obstructed at 6:45 pm and the photos taken in fieldwork show that the sun is obstructed at 6:46 pm. We further compared the predicted sun paths and the actual sun positions at 7 other locations. The validation results show that the proposed method can estimate the exact time point of sun glare been blocked with an error of less than 3 minutes, which further prove that the proposed method can estimate the sun glare with high accuracy.

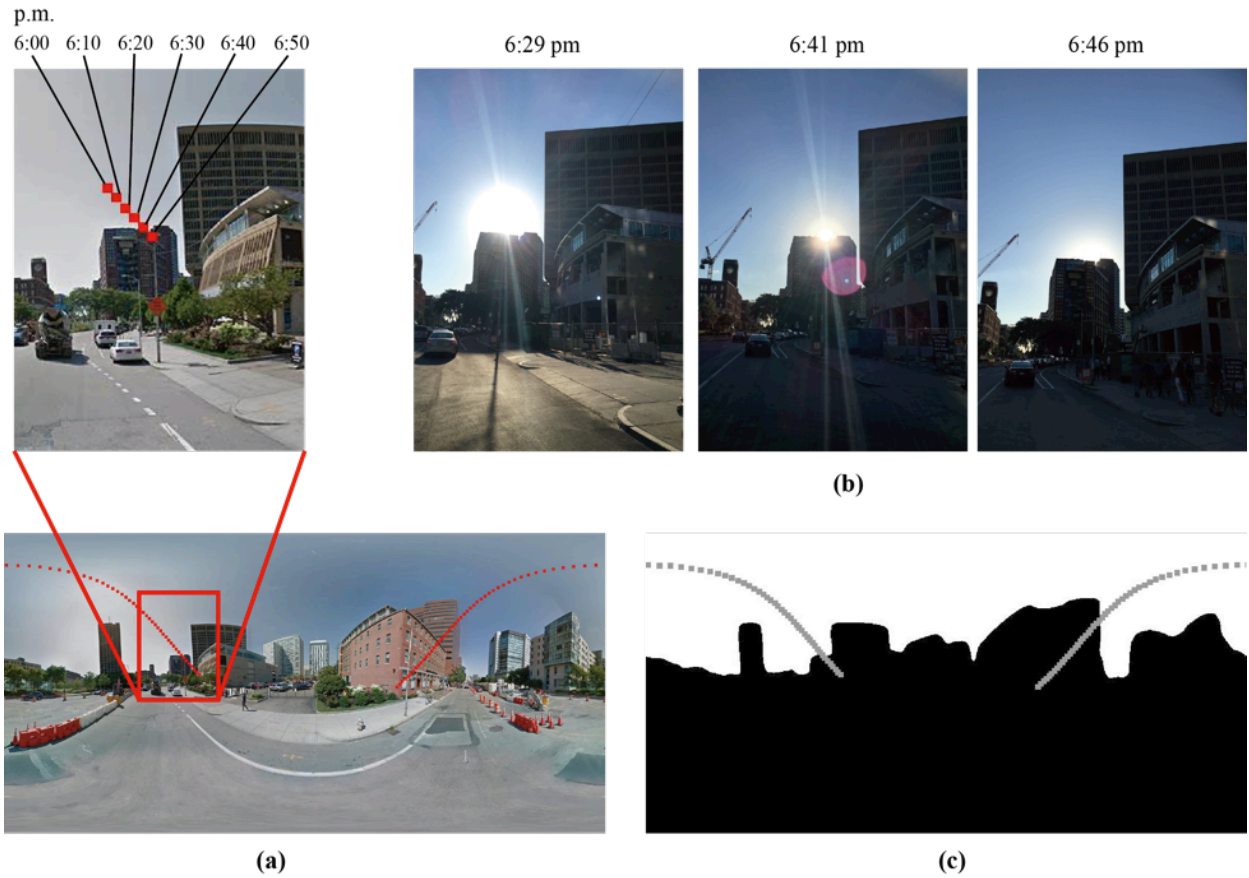

**Fig. 9.** The comparison of the predicted sun path on a GSV panorama and three photos taken at the same location at three different time points of July 5$^{th}$, 2018, (a) the sun path of July 5$^{th}$ on a GSV panorama, (b) three photos taken at the same location at different times, (c) the overlay of sun path of July 5$^{th}$ on the segmented GSV panorama, black part represents obstruction pixels and white part represent the sky pixels.

**Fig. 10** shows the spatial distributions of the sun glare in the study area on June 20$^{th}$ and September 20$^{th}$ with consideration of obstructions along the streets. Compared with the sun glare maps without considering the obstruction (**Fig. 6**), the existence of obstructions along the street sides reduces the occurrence and duration of sunrise and sunset sun glares significantly. On June 20th, these streets in the southern Cambridge, especially the Memorial Drive, are exposed to sunrise sun glare more than the rest areas. Those streets in the northwestern and the eastern Cambridge have more potential sunset sun glare presence than other regions on June 20th.

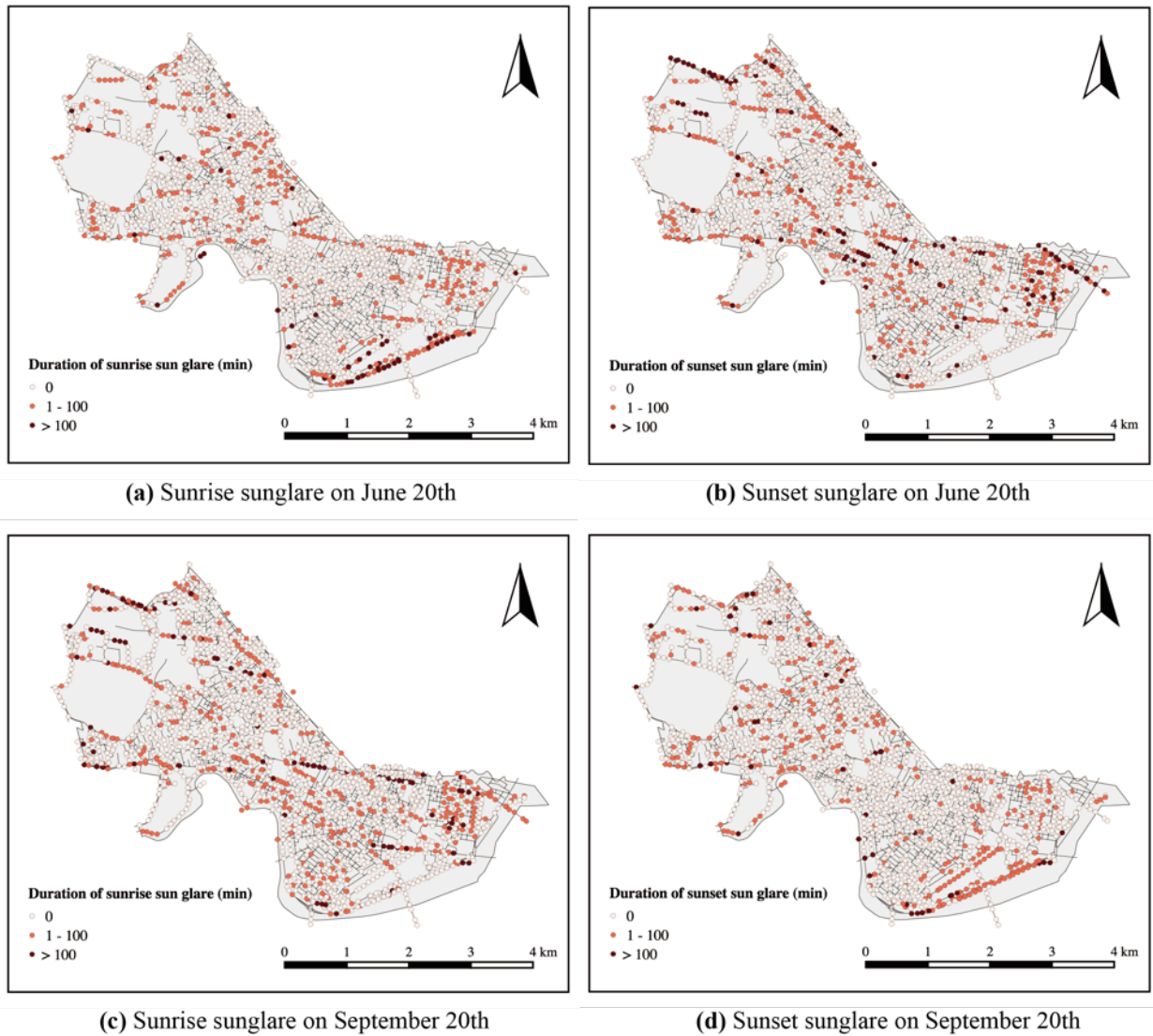

**Fig. 10.** The spatial distributions of sun glares in Cambridge, MA on June 20th and September 20th, 2018 with consideration of obstructions, (a) the spatial distribution of the sunrise glare on June 20th, (b) the spatial distribution of the sunset glare on June 20th, (c) the spatial distribution of sunrise glare on September 20th, (d) the spatial distribution of actual sunset glare on September 20th.

## 4. Discussion

Sun glare is one of the most annoying and dangerous hazards caused by weather for most drivers. It is important for drivers to be informed the potential presence of sun glare in order to avoid sun glare caused accidents. Based on the geometrical model of sun position and the locations of drivers, it is possible to predict the occurrence of sun glare by judging the relative angle with sun from the perspective of drivers. In order to predict the occurrence of sun glare, three major factors need to be considered, sun position, the driving direction, and the surrounding obstructions. Different from previous studies using digital terrain models to predict the sun glare, this study proposed to use the publicly accessible and globally available Google Street View (GSV) panoramas to predict and map the presence of sunrise and sunset sun glares. Considering the fact that GSV images were captured by vehicles on driveways, therefore, the GSV represents what drivers can see on driveways, which would make GSV based method more suitable to predict the sun glare. The presence of obstruction of sun glare at one location of a specific time can be judged by overlaying the sun position on the corresponding GSV panorama and checking whether the sun is located on obstruction or sky pixels in segmented GSV panorama. Therefore, the proposed GSV based sun glare prediction method can predict the presence of the sun glare more accurately and objectively.

The proposed method was applied to Cambridge, Massachusetts to map the spatio-temporal distributions of sunrise and sunset sun glares. Results show that the sun glare distributions change significantly in different months of one year. The orientation ranges of streets that potentially have sun glares change monthly, because of the changing sun paths in different months. Validation results show that the method can help to predict the sun glare precisely. Theoretically, the method can generate the presence of the sun glare map at day level. In this study, for illustration purpose, we presented the sun glare maps in two days of different months in one year.

The proposed method is totally automatic, and without any human intervention. Therefore, the proposed method provides a simple way to predict the sun glare on driveways at large scales. In addition, the study is based on road networks and the publicly accessible GSV images. Therefore, the method can be applied to any place with GSV service available. This would be very important for those areas with no high spatial resolution spatial data (digital terrain model and building height model, or digital surface model) available.

Previous studies have used the digital terrain models to predict the presence of the sun glare. However, the digital terrain models usually have very low spatial resolutions, and buildings and trees, which would obstruct the sun glares, are usually not considered. Different from the digital terrain model based model for sun glare prediction, the proposed GSV based method applies the geometrical model of the sun and cars to cylindrical GSV panoramas to predict the obstruction of sun glares at any given time for any given location from driver's perspective. The GSV based method considers the influence of fine-level obstructions on roadsides, such as trees and building blocks.

The proposed method for sun glare prediction would help drivers to be informed about the potential sun glares while driving. For navigators and auto insurance companies, the sun glare map can help them to notify the potential sun glare presences to drivers and customers, which would help drivers and customers to get a better and safer driving experience. The proposed method for sun glare prediction can also provide useful information for the Department of Transportation. The maintainers of roads can zone those high risky areas and take measure to minimize the potential negative impacts of sun glare on drivers.

There are still some limitations in this study. The seasonal change of the street greenery would also influence the presence of the sun glare considering the fact that street greenery cannot block the sun glare in leaf-off season. In this study, we only generated the sun glare maps during leaf-on season since the unavailability of GSV images taken in leaf-off season. Future studies should also generate the sun glare presence map for leaf-off season, when the street-level images are available for leaf-off season. Secondly, current method only considers the occurrence of the sun glare and the intensity of the sun glare is not considered. Future studies should also consider the intensities of sun glares and the impacts of different sun glare intensities on people of different age-gender groups. In addition, more studies need to be done about how the sun glare influencing the driving experience and the traffic accidents at a large scale.

## 5. Conclusions

This study proves that it is possible to use Google Street View images to predict and map the occurrence of the sun glare accurately. The proposed method in this study would help to generate the spatio-temporal distribution of the occurrence of sun glare, which would help drivers, cyclists, and pedestrians to minimize the negative impacts of sun glare and decrease the potential traffic accidents caused by sun glares. The generated spatio-temporal distribution of sun glare would also help policy makers to redesign road infrastructures to curb the influence of sun glare on drivers, cyclists, and pedestrians. Considering the public accessibility and global availability of street-level images, this study would bring a general way to predict the sun glare and forecast the sun glare for anywhere with street-level images available.